%% file: main.tex
\documentclass[preprint,12pt]{elsarticle}

\usepackage{amssymb}

\usepackage{amsmath}
\usepackage{dsfont}
\usepackage{algorithm}
\usepackage{algorithmic}
\usepackage{subfigure}
\usepackage{graphicx}
\usepackage{wrapfig}
\usepackage{multirow}
\usepackage{xcolor}
\usepackage{colortbl}
\usepackage{booktabs}
\usepackage{url}

\definecolor{c_gray}{gray}{.9}

\newcommand{\modified}[1]{{\color{black}{#1}}}

\usepackage{rotating}

\begin{document}

\begin{frontmatter}

\title{NAS-BNN: Neural Architecture Search for \\ Binary Neural Networks}

\author[label1]{Zhihao Lin}
\ead{linzhihao@stu.pku.edu.cn}

\author[label1]{Yongtao Wang\corref{cor1}}
\ead{wyt@pku.edu.cn}

\author[label1]{Jinhe Zhang}
\ead{jinhezhang17@gmail.com}

\author[label1]{Xiaojie Chu}
\ead{chuxiaojie@stu.pku.edu.cn}

\author[label2]{Haibin Ling}
\ead{hling@cs.stonybrook.edu}

\cortext[cor1]{Corresponding author}

\affiliation[label1]{organization={Wangxuan Institute of Computer Technology, Peking University},
            addressline={128 North Zhongguancun Road},
            city={Beijing},
            postcode={100080},
            country={China}}

\affiliation[label2]{organization={Department of Computer Science, Stony Brook University},
            addressline={100 Nicolls Road},
            city={Stony Brook},
            postcode={NY 11794},
            state={New York},
            country={USA}}

\input{sections/0_abstract.tex}

\begin{keyword}
Neural architecture search \sep Binary neural network \sep Deep learning
\end{keyword}

\end{frontmatter}

\input{sections/1_introduction}

\input{sections/2_related_work}

\input{sections/3_methodology}

\input{sections/4_experiments}

\input{sections/5_conclusion}

\input{sections/6_reference}

\end{document}

%% file: sections/0_abstract.tex
\begin{abstract}

Binary Neural Networks~(BNNs) have gained extensive attention for their superior inferencing efficiency and compression ratio compared to traditional full-precision networks. \modified{However, due to the unique characteristics of BNNs, designing a powerful binary architecture is challenging and often requires significant manpower.} A promising solution is to utilize Neural Architecture Search~(NAS) to assist in designing BNNs, \modified{but current NAS methods for BNNs are relatively straightforward and leave a performance gap between the searched models and manually designed ones.} To address this gap, we propose a novel neural architecture search scheme for binary neural networks, named NAS-BNN. We first carefully design a search space based on the unique characteristics of BNNs. Then, we present three training strategies, which significantly enhance the training of supernet and boost the performance of all subnets. Our discovered binary model family outperforms previous BNNs for a wide range of operations~(OPs) from 20M to 200M. For instance, we achieve 68.20\% top-1 accuracy on ImageNet with only 57M OPs. \modified{In addition, we validate the transferability of these searched BNNs on the object detection task, and our binary detectors with the searched BNNs achieve a novel state-of-the-art result, \emph{e.g.}, 31.6\% mAP with 370M OPs, on MS COCO dataset. The source code and models will be released at \url{https://github.com/VDIGPKU/NAS-BNN}.}

\end{abstract}

%% file: sections/1_introduction.tex
\section{Introduction}

Binary Neural Networks~(BNNs)~\cite{qin2020binary}, whose weights and activations are both binary(1-bit), have gained popularity in various computer vision~(CV) tasks, such as image classification~\cite{hubara2016binarized,liu2018bi,ignatov2020controlling,phan2020mobinet,liu2020reactnet}, object detection~\cite{peng2019bdnn, wang2020bidet,zhao2022data,xu2021layer}, and point cloud applications~\cite{qin2020bipointnet}. As the binary convolution can be implemented by XNOR and bit-count operations, BNNs are able to achieve at least 58$\times$ speedup in CPUs or embedded devices~\cite{rastegari2016xnor, jiang2019bitstream} compared to full-precision networks.
However, due to the inaccurate gradients and lower fitting ability, the training phase of BNNs requires more iterations than full-precision networks to reach convergence. Additionally, the training of BNNs must be simulated with full precision rather than binary on GPUs, resulting in 64$\times$ more calculations than the theoretical operations~(OPs). These challenges make real-world deployments on diverse application scenarios difficult, as multiple models must be designed and then trained and validated through trial and error.

To address the aforementioned challenges, researchers leverage Neural Architecture Search~(NAS) to automatically design a series of BNNs~\cite{kim2020learning,bulat2020bats,zhu2020nasb,zhao2020bars,phan2020binarizing}. For example, BNAS~\cite{kim2020learning} and BATS~\cite{bulat2020bats} employ differentiable NAS~\cite{liu2018darts} on their carefully designed cell-based search spaces. Binary MobileNet~\cite{phan2020binarizing} uses weight-sharing NAS to explore the best candidate for the number of groups and find a tiny yet efficient BNN. \modified{Although these methods explore NAS methods for BNNs, they are all relatively straightforward, and the searched models are no longer competitive with manually designed ones.
In contrast, full-precision NAS~\cite{cai2019once,yu2019universally,yu2020bignas,wang2021attentivenas,wang2021alphanet,hu2022learning,tong2022neural} has been rapidly developed, and many effective methods have been proposed.
However, all of these methods focus on full-precision networks, and it is challenging to adopt them on BNNs directly, due to the main difficulties as follows:}

\textbf{Model architecture.}
Previous works have shown significant differences in the architecture design between BNNs and full-precision networks, and have revealed that one cannot simply adopt the search space of full-precision networks~\cite{kim2020learning,bulat2020bats} for BNNs. For example, Bi-Real Net~\cite{liu2018bi} introduces a \emph{Bi-Real} module to propagate the real-valued features through a shortcut and demonstrates that it can significantly improve the performance of BNNs. However, this module is useless and redundant in full-precision networks. Additionally, BNAS~\cite{kim2020learning} discovers that the \emph{Zeroise} layer, which is only used as a placeholder in full-precision NAS, can improve the accuracy of BNNs.

\textbf{Training strategy.}
The training strategies for BNNs and full-precision networks differ significantly. For instance, full-precision classification networks are often trained from scratch~\cite{he2016deep,howard2017mobilenets}, while BNNs typically rely on recently proposed two-step training process~\cite{bulat2019improved,martinez2020training}, which requires an extra pre-training process. Therefore, it is important to explore specific NAS training strategies that are tailored for BNNs.

To deal with the above difficulties, we present NAS-BNN, \modified{with a novel search space for BNNs and a set of training strategies that are specifically tailored to binary supernet.} We begin by evaluating the strengths and weaknesses of various model templates in BNNs, which leads us to develop our search space based on MobileNetV1~\cite{howard2017mobilenets}. Consistent with MoBiNet~\cite{phan2020mobinet}, we replace all \emph{depth-wise} convolution layers with normal convolution to reduce information loss in BNNs. We then incorporate the number of groups on each convolution into our search space to take full advantage of \emph{separable} convolution and to identify optimal architectures. Subsequently, we propose Non-Decreasing~(ND) constraint to eliminate ineffective subnets and obtain the final search space, \modified{which ensures each subnet retains as much information as possible.}
Then, we revisit the sandwich rule~\cite{yu2019universally, yu2020bignas} and introduce Bi-Teacher to guide the optimization process, achieving better convergence for all subnets. We also explore the optimization of shared weights by introducing Bi-Transformation, a learnable transformation that transfers shared weights from full-precision domain to binary domain. Additionally, we propose channel-wise weight normalization to maximize the information capacity of binary weights. Finally, we investigate the balance of supernet training and subnet finetuning, and propose two pipelines for model deployment.
\modified{Through these methods, we significantly improve the convergence of the supernet compared with the baseline, and achieve better accuracy-efficiency trade-offs on BNNs.}

Our main contributions are four-fold:
\begin{itemize}
    \item We design a novel search space for BNNs, including the traditional configurations~(depth, channel width, kernel size) and the specific configurations for BNNs~(the number of groups). In addition, we propose Non-Decreasing~(ND) constraint to remove the ineffective subnets, to build a more powerful search space.
    \item We introduce three training techniques to enhance the binary supernet, so as to boost the performance of each subnet.
    \item We present two deployment pipelines that aim to balance supernet training and subnet finetuning.
    \item Our binary model family, NAS-BNN, outperforms prior BNNs on a wide range of operations~(OPs). Notably, we achieve 68.20\% top-1 accuracy on ImageNet dataset with only 57M OPs for classification task, \modified{and 31.6\% mAP on MS COCO dataset with 370M OPs for object detection task.}
\end{itemize}

%% file: sections/2_related_work.tex
\section{Related work}
\subsection{Binary neural network}
Binary Neural Networks~(BNNs)~\cite{hubara2016binarized, rastegari2016xnor, liu2018bi, bethge2020meliusnet, liu2020reactnet} are the most extreme case of network quantization~\cite{esser2019learned,lin2022fq}, in which both weights and activations are in 1-bit. Binarized Neural Network~\cite{hubara2016binarized} first restricts both weights and activations to +1 and -1 and achieves nearly state-of-the-art results on small datasets~(\emph{i.e.}, MNIST). 
XNOR-Net~\cite{rastegari2016xnor} uses the sign function with an extra scaling factor to binarize the weights, which effectively reduces quantization error. 
However, these methods inevitably suffer from accuracy degradation on large-scale datasets such as ImageNet.
Subsequent researches~\cite{liu2018bi,phan2020mobinet,bethge2020meliusnet,liu2020reactnet} have shown that the poor performance of these BNNs on large-scale datasets is attributable to the inadequate model architecture. To address this issue, many methods have been proposed to design suitable architectures that reduce the performance gap between binary and full-precision networks.
Bi-Real Net~\cite{liu2018bi} introduces additional shortcuts to BNNs, significantly improving their performance.
MoBiNet~\cite{phan2020mobinet} proposes the $K$-dependency strategy to divide input channels into different groups, achieving a better trade-off between accuracy and efficiency.
MeliusNet~\cite{bethge2020meliusnet} presents Dense Block and Improvement Block to enhance model performance, achieving the accuracy of full-precision MobileNetV1~\cite{howard2017mobilenets} for the first time.
ReActNet~\cite{liu2020reactnet} adds additional learnable parameters to adjust the distribution of features and achieves state-of-the-art performance on BNNs. 

Although these methods have significantly improved the performance of BNNs, their architectures are manually designed and may be sub-optimal. This motivates us to explore the NAS method for BNNs.

\subsection{Neural architecture search}
Early NAS methods~\cite{zoph2016neural,liu2017hierarchical,zoph2018learning,real2019regularized} discover the optimal architecture by training and evaluating thousands of candidates from scratch, which is computationally expensive. Later, weight-sharing NAS methods~\cite{brock2017smash,pham2018efficient,cai2018proxylessnas,wu2019fbnet,guo2020single} are proposed to reduce the search cost. These methods train a weight-sharing supernet to rank the performance of subnets and then use a search algorithm to find the optimal architecture or performance Pareto front. 
{From full-precision case to low-bit case, there are some studies~\cite{wang2020apq,bai2021batchquant} to use NAS with low-bit quantization, to search the optimal architecture and mixed-precision configuration. APQ~\cite{wang2020apq} jointly searches the neural architecture and the layer-wise quantization policy, unifying the conventionally separated stages into an integrated solution. BatchQuant~\cite{bai2021batchquant} proposes a robust quantizer for Quantized-for-all supernet, and is the first work to support one-shot weight-sharing supernet with arbitrary mixed-precision quantization policy without retraining. Despite these methods achieve impressive results, they all focus on solving the quantization problem (\emph{i.e.}, bit$\ge$2) rather than the binarization problem (\emph{i.e.}, bit=1).}
From low-bit case to binary case, there are some pioneer studies to introduce NAS to BNNs.
BNAS~\cite{kim2020learning} and BATS~\cite{bulat2020bats} use the cell-based searching method to search for the optimal cell block. 
However, the drawbacks of cell collapsing are significantly amplified in BNNs, which leads to an irreversible tendency to choose parameter-free operations. 
Binary MobileNet~\cite{phan2020binarizing} adopts the weight-sharing NAS method to search the group numbers of BNNs.
Despite these attempts, the performance of NAS-based BNNs still lags behind that of manually designed ones, which is markedly different from the full-precision case.

%% file: sections/3_methodology.tex
\section{Methodology}

In this section, we elaborate on our neural architecture search scheme for binary neural networks.
We first present the problem definition of our task. Then, we introduce our carefully designed search space and three training techniques for BNNs. Finally, we present two pipelines for model deployment.

\subsection{Problem definition}
In BNNs, weights and activations are binarized through a sign function, that is,
\begin{equation}
    x_b = {\operatorname{Sign}}(x_r) = \left\{  
   \begin{array}{cc}
   + 1, & \operatorname{if} \ x_r \geq 0, \\  
   - 1, & \operatorname{otherwise},
   \end{array}
    \right.
\end{equation}
where $x_r$ indicates the full-precision weights or activations and $x_b$ is the binary counterpart. With this binarization step, the computationally expensive matrix multiplication between weights and activations can be replaced by lightweight XNOR and bit-count operations, thus saving time and resources.

Assuming that the search space of our proposed NAS-BNN is $\mathcal{A}$ and the architecture of a subnet is $a\in\mathcal{A}$, we can formalize the problem as
\begin{equation}
    \max_{\mathcal{W}_r} \sum_{{a}\in\mathcal{A}} \operatorname{Acc}_{\rm val}(\mathcal{N}(a, \mathcal{W}_b)),
\end{equation}
where $\mathcal{W}_r, \mathcal{W}_b$ denote the full-precision and binary weights respectively, $\mathcal{N}$ denotes the sampled subnet with architecture $a$ and binary weights $\mathcal{W}_b$, \modified{and $\operatorname{Acc}_{\rm val}$ represents the accuracy of the input architecture on the validation set.} The objective is to optimize $\mathcal{W}_r$ to improve the accuracy of all sampled subnets in the search space $\mathcal{A}$.

\begin{figure*}[!t]
\centering
\includegraphics[width=0.95\textwidth]{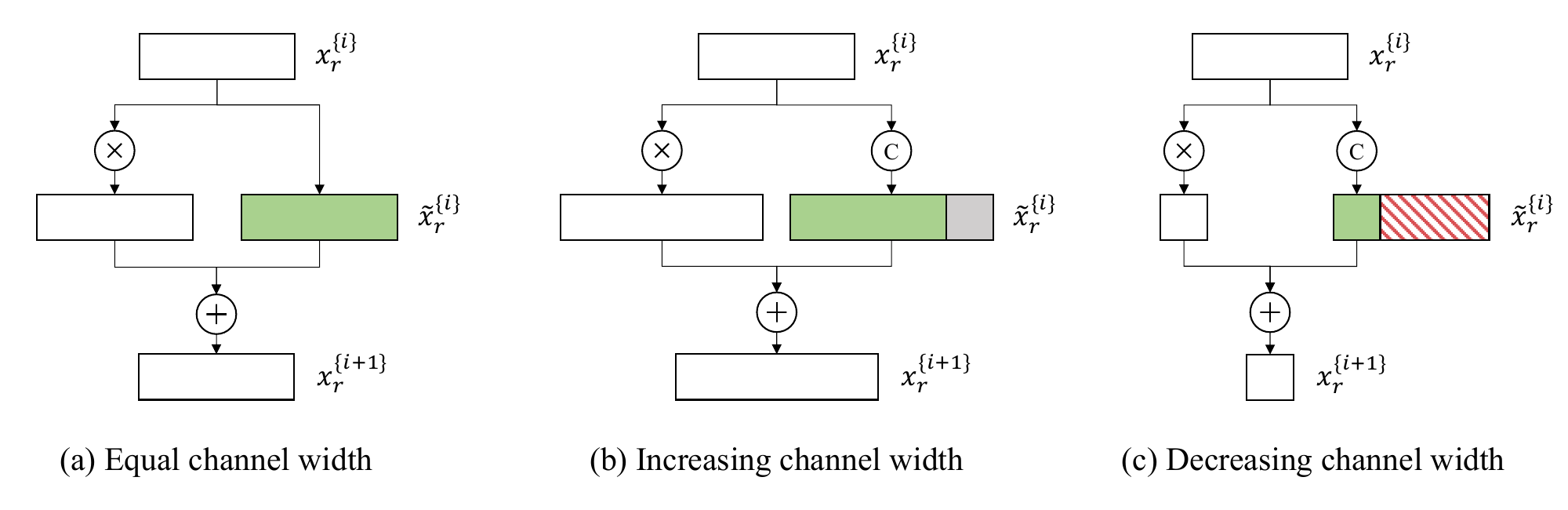}
\caption{Comparison of the cases with different channel widths in a \emph{Bi-Real} module. ``$\times$" denotes the binary convolution, ``\textrm{C}" denotes the tiling channel concatenation, and ``$+$" denotes the element-wise addition. (a) and (b) illustrate equal and increasing cases, respectively, while (c) illustrates the case where channel width decreases. It shows a significant loss of feature information in (c)~(\emph{i.e.}, the portion denoted by the slash).}
\label{fig.2}
\end{figure*}

\subsection{Search space}
In this section, we present our carefully designed search space for binary neural networks. To avoid the significant loss of feature information in \emph{Bi-Real} module~\cite{liu2018bi} (\emph{i.e.}, the critical component of BNNs) as shown in Fig.~\ref{fig.2}~(c), we do not choose those models with Bottleneck blocks~(\emph{e.g.}, ResNet~\cite{he2016deep}, MobileNetV2~\cite{sandler2018mobilenetv2}, \emph{etc.}). 
Thus, we design our search space based on the MobileNetV1~\cite{howard2017mobilenets} template with \emph{separable} convolution.
Generally, each \emph{separable} convolution contains one \emph{depth-wise} convolution and one \emph{point-wise} convolution. 
{However, the \emph{depth-wise} convolution is not effective in the binary case~\cite{kim2020learning,bulat2020bats}, since the output values are limited within a narrow range~\cite{phan2020mobinet}. For example, a binary $3\times3$ depth-wise filter convolving with one input channel of the input yields values in the range of $[-9, 9]$, which degrades the representation capability of the binary neural networks. On the other hand, binary normal convolution or group convolution results in a larger output value range, allowing for more abundant feature representation and effectively preserving the distribution of the data samples through network layers.} Therefore, to further reduce the loss of information, we replace all \emph{depth-wise} convolution layers with normal convolution and search for their number of groups to achieve the global optimum.

\input{sections/tables/search_space}

Our search space, which is summarized in Table~\ref{tab:search_space}, includes both traditional configurations~(depth, channel width, kernel size) and specific configurations for BNNs~(the number of groups). 
Similar to other works~(\emph{e.g.}, OFA~\cite{cai2019once}), each searchable unit can adopt an arbitrary number of layers, and each layer can have an arbitrary channel width, kernel size, and number of groups. 
The candidates for the depth are set based on MobileNetV1~\cite{howard2017mobilenets}, with similar depth ratio~(\emph{i.e.}, 1:1:4:1).
The candidates for the number of groups are set based on MoBiNet~\cite{phan2020mobinet}, ensuring that each group contains at least 48 dependent channels to maintain a wide value range for the feature representations~\cite{phan2020binarizing}.
Moreover, we only search the first convolution layer in each \emph{separable} convolution, while the second one is fixed as a normal 1x1 convolution, as it is primarily utilized to fuse information between different channels.
Overall, that search space contains approximately $4.0\times10^{21}$ candidate architectures, and all of them can inherit the shared weights from one supernet.

\textbf{Non-Decreasing constraint.}
Although the search space presented above avoids the usage of Bottleneck block, it still contains many sub-optimal networks that have adjacent convolutions with decreasing feature channels.
Therefore, we present the Non-Decreasing~(ND) constraint to remove subnets with decreasing channel width from our search space. The final search space $\mathcal{\widetilde A}$ can be formulated as
\begin{equation}
    \mathcal{\widetilde A} = \{a\mid c_{a}^{\{i+1\}}\geq c_{a}^{\{i\}}, 1\leq i\leq L-1, \ a\in\mathcal{A}\},
\end{equation}
where $L$ denotes the total number of layers, and $c_{a}^{\{i\}}$ denotes the channel width of layer $i$ with the architecture $a$.

With this constraint, we remove the sub-optimal subnets and also shrink the search space. Specifically, the size of the search space is reduced to 0.01\% of the original one, a total of $3.3\times10^{17}$ subnets. As a result, the training and searching phases can be significantly accelerated.

\subsection{Training for binary supernet}
In this section, we introduce three simple yet efficient training techniques~(\emph{i.e.}, Bi-Teacher, Bi-Transformation, and Weight Normalization) to obtain a better binary supernet, so as to boost the performance of each subnet. These techniques are inspired by the unique characteristics of BNNs, which are distinct from previous NAS methods for full-precision networks.

\subsubsection{Bi-Teacher for sandwich rule}
The sandwich rule~\cite{yu2019universally} has demonstrated excellent convergence behavior and overall performance in previous NAS methods. According to this rule, the largest, smallest, and two random subnets are trained at each iteration. Additionally, the soft label predicted by the largest network are retained as supervision information to guide the training of the smallest and random subnets, referred to as inplace distillation~\cite{yu2019universally}.
There are two primary reasons for selecting the largest network as the teacher in NAS methods. Firstly, it represents the upper bound of the performance of all subnets in search space, and training it at each iteration can effectively enhance the performance of all subnets. Secondly, it is the most potent network of the supernet, which can best guide the training of each subnet and hasten the convergence process.
However, the choice of the teacher for binary NAS is not straightforward. If we consider the first reason, we will be inclined to select the largest Binary Weights and Binary Activations~\emph{(BWBA)} network. On the other hand, if we consider the second reason, we will select the largest Full-precision Weights and Full-precision Activations~\emph{(FWFA)} network, leading to a contradiction.

Different from the above two choices, we introduce a trade-off scheme that adopts the largest Full-precision Weights and Binary Activations~\emph{(FWBA)} network as the teacher, named Bi-Teacher. 
{We refrain from using binary weights (\emph{i.e.}, \emph{BWBA}) because the gradient error incurred by binarization is substantial, causing a ``gradient mismatch'' for all weights and leading to bad convergence. Meanwhile, we also refrain from using full-precision activations (\emph{i.e.}, \emph{FWFA}) to  mitigate the collapse of the optimal space to the full-precision domain.
Overall, our Bi-Teacher benefits from the full-precision weights and binary activations, which can accelerate the convergence of the supernet and prevent the collapse problem.}

The overall training loss function can be formulated as
\begin{equation}
    \mathcal{L}= \mathcal{L}_{\rm CE}(\hat{y}_l, Y) + \mathcal{L}_{\rm KL}(y_s, \hat{y}_l) +
    \sum_{i} \mathcal{L}_{\rm KL}(y_{r_i}, \hat{y}_l),
\end{equation}
where $\mathcal{L}_{\rm CE}$ and $\mathcal{L}_{\rm KL}$ denote the cross-entropy loss and the KL divergence loss, respectively, $Y$ denotes the ground truth label, $\hat{y}_l$ denotes the predictions of Bi-Teacher, and $y_{s},y_{r_i}$ denote the predictions of the smallest and random binary subnets, respectively.

\begin{figure*}[!t]
\centering
\includegraphics[width=0.95\textwidth]{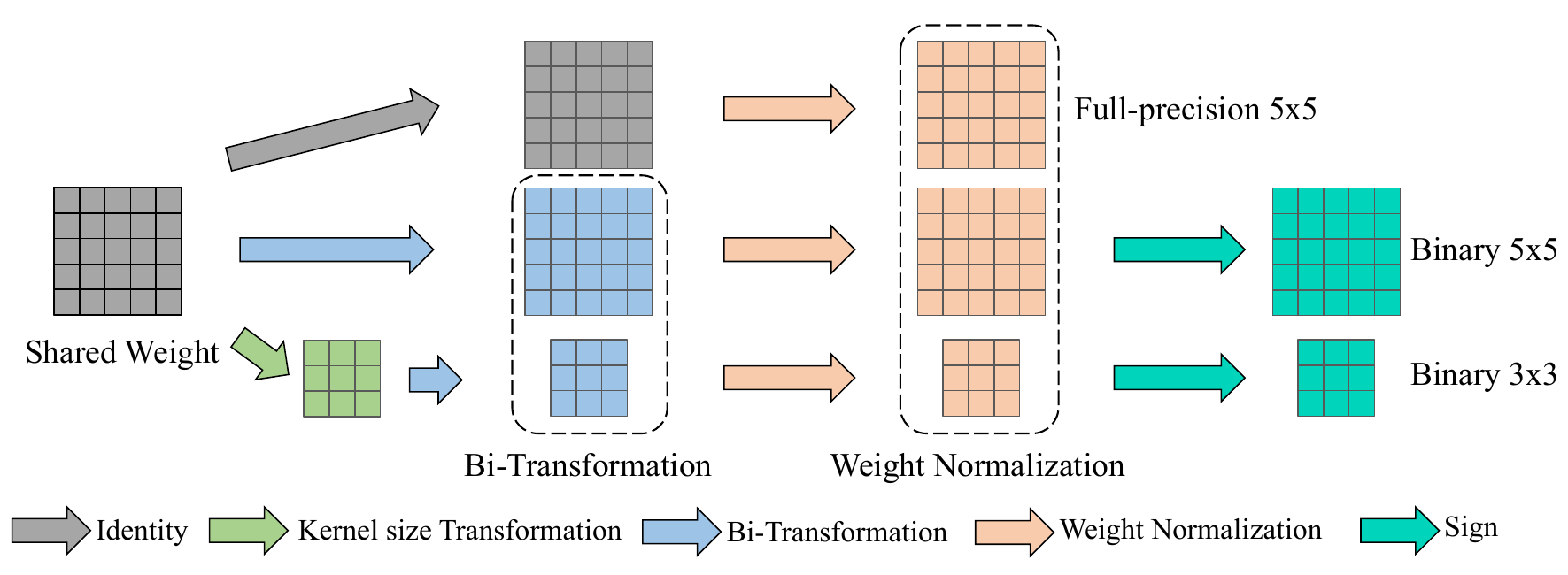}
\caption{Illustration of the process to obtain full-precision and binary weights from a shared weight.}
\label{fig.3}
\end{figure*}

\subsubsection{Bi-Transformation for shared weights}
The aforementioned Bi-Teacher utilizes the full-precision weights to enhance the convergence of all subnets, while the ultimately desired subnets are in binary. In that case, the shared weights have two different roles as they must fit both full-precision and binary domains. To address this issue, as shown in Fig.~\ref{fig.3}, we introduce an extra learnable transformation, Bi-Transformation, for shared weights to transfer them from full-precision to binary domain. 
Taking the conversion from $k\times k$ full-precision domain weight ($\mathcal{W}_r$) to the binary domain ($\mathcal{W}_b$) as an example, we introduce a learnable transformation matrix $\theta\in \mathbb{R}^{k^2\times k^2}$ and initialize it as a diagonal matrix. Then, we use matrix multiplication to obtain the binary domain's weight as
\begin{equation}
    \mathcal{W}_{b}= \operatorname{Reshape}_{k\times k}(\operatorname{Flatten}(\mathcal{W}_{r})\cdot\theta).
\end{equation}

Following OFA~\cite{cai2019once}, we adopt different transformation matrices for different layers, while using one shared matrix for all channels within each layer. Therefore, the additional parameters introduced by Bi-Transformation can be negligible.

\subsubsection{Weight Normalization}
To further enhance the information capacity of subnets, as shown in Fig.~\ref{fig.3}, we introduce channel-wise weight normalization for each weight $\mathcal{W}$ as
\begin{equation}
    \mathcal{W}^{'}= \frac{\mathcal{W} - \mu_{\mathcal{W}}}{\sqrt{\sigma_{\mathcal{W}}^2 + 10^{-5}}}, 
\end{equation}
where $\mu_{\mathcal{W}}$ and $\sigma_{\mathcal{W}}$ denote the channel-wise mean and standard deviation of weight $\mathcal{W}$, respectively.

{This simple yet efficient technique is widely used in BNN studies~\cite{Li2020Additive, qin2020forward}. It ensures that each weight has the zero-mean attribute and that the counts of $+1$ and $-1$ in one binary weight are as consistent as possible, which makes binary weights achieve maximum information entropy, leading a stronger representation ability. Meanwhile, introducing weight normalization to weight-sharing NAS can also decouple the shared weights in representation, making the shared parameters more flexible for all subnets.}

\subsection{Deployment pipelines}
Unlike full-precision NAS methods~\cite{wang2021attentivenas,wang2021alphanet}, our framework comprises of two parallel deployment pipelines. \textbf{Pipeline 1} is similar to these methods and to deploy the subnets with inherited parameters after the training and searching steps. \textbf{Pipeline 2} introduces an additional finetuning step for the searched subnets based on pipeline 1. The main motivation for pipeline 2 is that the training of BNNs has to be simulated with full precision on GPUs rather than binary, resulting in 64$\times$ more calculations than the theoretical operations~(OPs). This makes it impossible to pursue the full convergence of the supernet with limited computing resources. Therefore, in this case, we believe that the pursuit of the balance of supernet training and subnet finetuning is more appropriate.
On the whole, pipeline 1 only requires one training to obtain many workable subnets, while pipeline 2 can further improve the performance of these subnets.

%% file: sections/tables/search_space.tex
\begin{table*}[!t]
\centering
\caption{The proposed binary search space. ``k'' represents the number of classes.}
\label{tab:search_space}
\vspace{10pt}
\small
\resizebox{\linewidth}{!}{
\begin{tabular}{ccccccc}
\toprule
Input                  & Operator       & Depth    & Channel width       & Kernel size & Groups    & Stride \\
\midrule
224 $\times$ 224        & Conv2d           & 1        & \{24, 32, 48\}      & 3           & 1         & 2      \\
112 $\times$ 112        & Separable Conv2d & \{2, 3\} & \{48, 64, 96\}      & 3           & 1         & 1      \\
112 $\times$ 112        & Separable Conv2d & \{2, 3\} & \{96, 128, 192\}    & \{3, 5\}    & \{1, 2\}  & 2      \\
56 $\times$ 56          & Separable Conv2d & \{2, 3\} & \{192, 256, 384\}   & \{3, 5\}    & \{2, 4\}  & 2      \\
28 $\times$ 28          & Separable Conv2d & \{8, 9\} & \{384, 512, 768\}   & \{3, 5\}    & \{4, 8\}  & 2      \\
14 $\times$ 14          & Separable Conv2d & \{2, 3\} & \{768, 1024, 1536\} & \{3, 5\}    & \{8, 16\} & 2      \\
7 $\times$ 7            & Avgpool2d        & 1        & -                   & -           & -         & -      \\
1 $\times$ 1            & Linear         & 1        & k                   & -           & -         & -      \\
\bottomrule
\end{tabular}
}
\end{table*}

%% file: sections/4_experiments.tex
\section{Experiments}

In this section, we present the experimental results of our proposed NAS-BNN. We first describe the detailed experimental setups. \modified{Then, we present the main results on ImageNet dataset and MS COCO dataset.}
After that, we conduct ablation studies to evaluate the effectiveness of each component.
In the end, we compare our method with other mixed-precision quantization methods.

\subsection{Experimental setups}

\subsubsection{Dataset}
The experiments of image classification are conducted on the ImageNet dataset. For the main results, we use ImageNet-1K, which consists of 1.2M training samples and 50K evaluation samples. For the ablation studies, we use ImageNet-100, a subset of ImageNet-1K, to quickly verify the effectiveness of each component. Regarding the validation set, we randomly sample 50 training images from each category and use them to evaluate the performance of subnets during the searching step.

\modified{The experiments of object detection are conducted on the widely used MS COCO dataset. Specifically, we use MS COCO 2017, which consists of 120K training samples and 5K evaluation samples.}

\subsubsection{Implementation details}
We conduct all experiments on 8 V100 GPUs. We adopt the binarization method of ReActNet~\cite{liu2020reactnet}. For the supernet training, Adam optimizer with the cosine learning rate decay scheduler is used. The initial learning rate is set to $5\times 10^{-4}$ and the weight decay is set to $5\times 10^{-6}$. We train the supernet for 512 epochs with a batch size of 512. In the searching step, we use the evolutionary algorithm to search the Pareto front. For the finetuning step of pipeline 2, the initial learning rate is set to $1\times10^{-5}$. \modified{For the training of object detector, we directly inherit the NAS-BNN model architectures searched on the ImageNet-1K dataset and use the shared weights from supernet as the initialization for the backbone network. We select Faster R-CNN as the binary detector framework.}
We quantize the first convolution layer to 8-bit with LSQ~\cite{esser2019learned} rather than keeping it at full precision, to further reduce the inference complexity of the final models. The total number of operations~(OPs) is calculated by (FLOPs + (Int8OPs / 8) + (BOPs / 64)), which is similar to \cite{liu2018bi,liu2020reactnet}.

\input{sections/tables/results}

\subsection{Main results on ImageNet}

As shown in Table~\ref{tab:results}, we summarize the performance of our binary model family, NAS-BNN, and compare it with several state-of-the-art BNNs, including those designed manually and automatically. For the models directly inheriting parameters from the supernet~(\emph{i.e.}, with pipeline 1), it is evident that our methods outperform all automatically designed~(\emph{i.e.}, NAS-based) models and most manually designed models. For instance, NAS-BNN-F achieves 69.54\% top-1 accuracy with 180M OPs.
We further evaluate the top-1 accuracy of the finetuned models~(\emph{i.e.}, with pipeline 2). We finetune each model with 100 epochs, which is only about 1/5 of the training cost for ReActNet. The results demonstrate that it significantly enhances the top-1 accuracy by more than 1\%. For instance, NAS-BNN-C \#100 achieves 69.49\% top-1 accuracy with 87M OPs, and NAS-BNN-F \#100 achieves 70.80\% top-1 accuracy with 180M OPs. 

Our total search cost on ImageNet is 240 GPU days~(220 GPU days for supernet training, and 20 GPU days for searching). This cost is acceptable when compared with the training cost of manually designed BNNs. For example, training a ReActNet-A needs 50 GPU days. Moreover, our NAS-BNN can directly inherit shared parameters to obtain a series of Pareto front subnets, making our models more flexible for diverse hardware platforms. We report the total cost~(\emph{i.e.}, search epochs and train/finetune epochs) for all models in Table~\ref{tab:results} instead of search cost since almost all NAS-based BNN methods do not report their search cost.
As we can see, the total cost of our method is significantly lower than that of other methods, especially when considering multiple deployment scenarios.
Last, it is worth noting that our method does not require an additional full-precision teacher~(\emph{e.g.}, ResNet34~\cite{he2016deep}) to supervise the training, which is an indispensable component for ReActNet.

\modified{
\subsection{Main results on MS COCO}
\input{sections/tables/results_det}
To validate the transferability of the architectures searched on the ImageNet image classification dataset, \emph{i.e.}, NAS-BNN-(A-F), we apply them to the object detection task by using them as the backbones and propose a series of binary detectors.
As shown in Table~\ref{tab:results_det}, we summarize the performance of our binary NAS-BNN detectors, and compare them with several state-of-the-art binary detectors. We can find that our models outperforms existing binary detectors across various computational budgets. For example, with less than 436M OPs, our model NAS-BNN-D achieves the highest mAP of 31.6\%; with less than 459M OPs, our model NAS-BNN-E also achieves the highest mAP of 32.1\%. Additionally, our proposed models expand the computational range of binary detectors by proposing an extremely small model with 140M OPs and a larger model with 484M OPs.
}

\subsection{Ablation studies}

\input{sections/tables/template}

\subsubsection{Effect of search space template}
To demonstrate the advantages of our search space template, we binarize three widely used models~(\emph{i.e.}, MobileNetV1~\cite{howard2017mobilenets}, MobileNetV2~\cite{sandler2018mobilenetv2} and ResNet50~\cite{he2016deep}) with \emph{Bi-Real} module~\cite{liu2018bi}, and evaluate their performance on the ImageNet-100 dataset. As per the previous works~\cite{liu2020reactnet,kim2020learning}, we replace all \emph{depth-wise} convolution layers in MobileNet with normal convolution.
The experimental results are presented in Table~\ref{tab:template}. It is evident that the binary MobileNetV1 outperforms other models in terms of both operations~(OPs) and accuracy.

\begin{figure*}[t]
\centering
\includegraphics[width=0.95\textwidth]{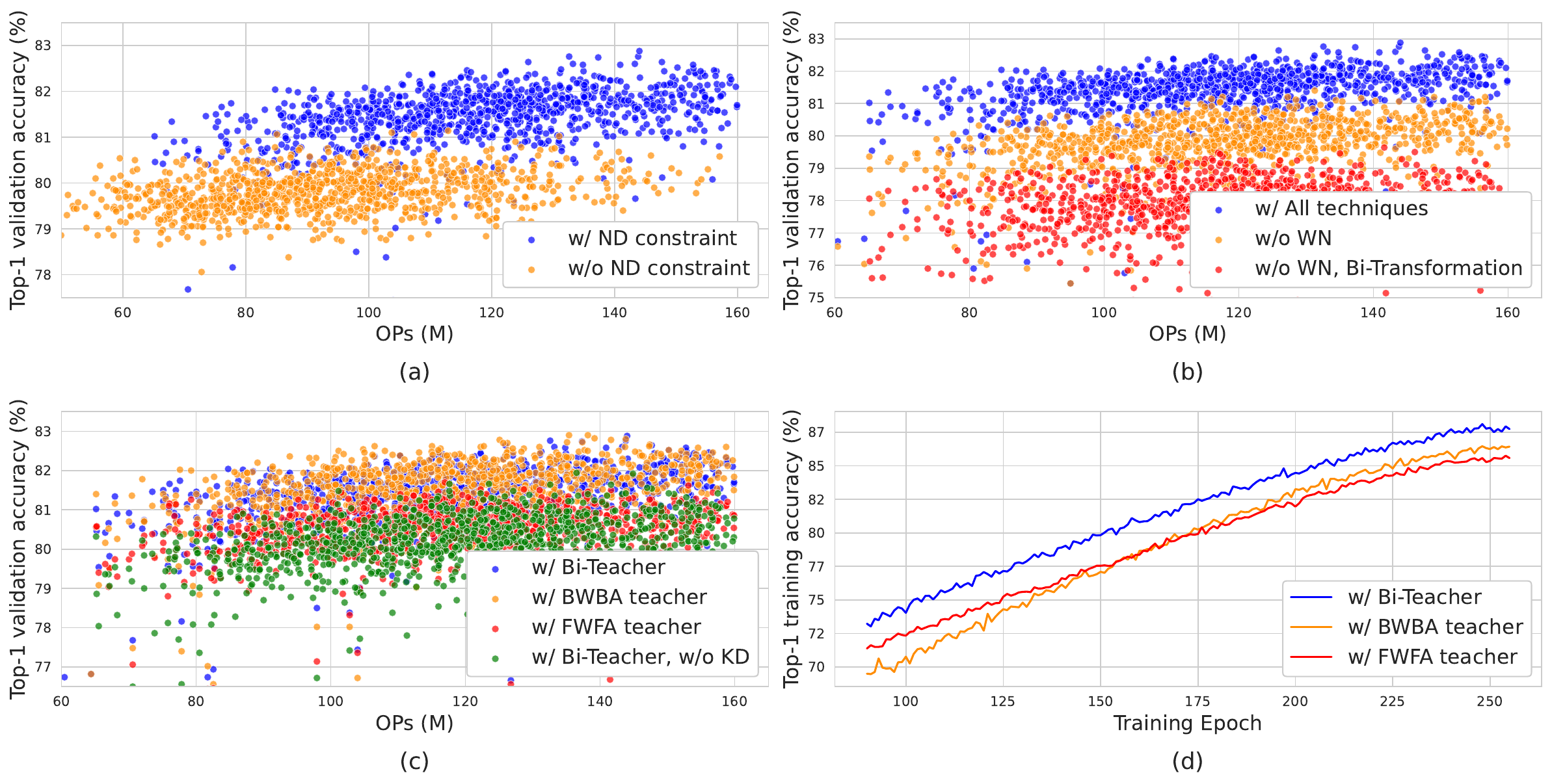}
\caption{Impact of the proposed training techniques on ImageNet-100. (a-c) We randomly sample 1000 subnets and directly evaluate the validation accuracy with inherited parameters from the supernet. (d) The training accuracy of random subnets with different teachers.}
\label{fig.4}
\end{figure*}

\subsubsection{Effect of Non-Decreasing~(ND) constraint}
\label{sec:nd}
In this section, we investigate the effect of the ND constraint, a shrinking strategy used in our search space. We train the supernet with and without ND constraint, and then randomly sample 1000 subnets to evaluate their top-1 accuracy on ImageNet-100 by inheriting parameters of the supernet.
As shown in Fig.~\ref{fig.4}~(a), we observe that the overall accuracy of the sampled subnets decreases by $\sim$1.5\% without ND constraint. This finding highlights that the ND constraint effectively removes sub-optimal samples and significantly enhances the overall performance of subnets.

\subsubsection{Effect of Bi-Transformation and Weight Normalization}
We ablate the impact of Bi-Transformation and Weight Normalization and present the results in Fig.~\ref{fig.4}~(b). Firstly, we verify the effect of the more straightforward method, Weight Normalization, and observe a degradation of $\sim$1.5\% in accuracy when we remove it. Secondly, we evaluate the more critical technology, Bi-Transformation, and observe a more significant decrease of $\sim$2\% in accuracy.

\subsubsection{Effect of Bi-Teacher}
We further study the effect of Bi-Teacher. Consistent with the above ablation studies, we verify it by evaluating the accuracy of randomly sampled subnets. As shown in Fig.~\ref{fig.4}~(c), Bi-Teacher significantly outperforms the \emph{FWFA} teacher and performs equivalently to \emph{BWBA}. To further verify the performance of Bi-Teacher and the \emph{BWBA} teacher, we show the training accuracy of random subnets in Fig.~\ref{fig.4}~(d). We observe that Bi-Teacher performs much better than the \emph{BWBA} teacher, causing the subnets to achieve sufficient convergence. {Besides, in the case of ``w/ Bi-teacher, w/o KD'', it is evident that when knowledge distillation is removed from the Bi-Teacher and the ground truth label is used directly to train the smallest subnet and two random subnets, there is a significant decrease in the overall accuracy of the subnets by $\sim$1.5\%.}

\subsubsection{Effect of the finetuning step on pipeline 2}

\input{sections/tables/finetune}
We conduct a series of experiments to explore the effect of the finetuning step on pipeline 2. Take NAS-BNN-B as an example, we finetune it with 0, 25, 50, 75, and 100 epochs. As shown in Table~\ref{tab:finetune}, the top-1 accuracy of NAS-BNN-B increases significantly, even with only 25 epochs~($\uparrow$1.20\%).
The results demonstrate that the finetuning step on pipeline 2 can significantly improve the performance, achieving a better trade-off between supernet training and subnet finetuning.

\subsection{{Comparison to the mixed-precision quantization methods}}

{
To demonstrate the effectiveness of to search models on binary domain, we compare our NAS-BNN models with mixed-precision quantization models in terms of BitOPs and model size. Our NAS-BNN-F \#100 achieves 70.8\% top-1 accuracy on ImageNet-1K with only 180M OPs (equivalent to 11.5G BitOPs) and a size of 9.3 MB, while mixed-precision quantization methods achieve lower top-1 accuracy with larger BitOPs or model size. For example, DNAS~\cite{wu2018mixed} achieves 70.6\% top-1 accuracy with 25.7G BitOPs, SPOS~\cite{guo2020single} achieves 70.5\% top-1 accuracy with 24.3G BitOPs, and One-Shot MPS~\cite{koryakovskiy2023one} achieves 69.9\% top-1 accuracy with approximately 14.8G BitOPs and a size of 27.1 MB.
}

\modified{
\section{Discussion}

In this section, we show the detailed architectures of our specialized models.

\input{sections/tables/arch}

\subsection{{Detailed architectures of NAS-BNN}}
{
We show the detailed architectures of our specialized NAS-BNN models in Table~\ref{tab:arch}. 
Upon examining the discovered Pareto front subnets, we can draw several intriguing conclusions that may aid us in designing more effective search spaces or model architectures.
First, it is better to opt for a larger number of groups rather than a smaller channel width or depth, when there is a need to reduce the size or computational operations of a binary neural network. For instance, in the 4-th stage of NAS-BNN-B, the number of groups is predominantly 8, while the channel width remains almost unchanged compared to NAS-BNN-C.
Second, in deeper layers, it is recommended to increase the proportion of convolutions with a kernel size of 5. For example, in NAS-BNN-F, the proportion of convolution with a kernel size of 5 is 1/5 in 2-nd and 3-rd stages, but it rises to 2/3 in the 4-th stage.

}

}

%% file: sections/tables/results.tex
\begin{table*}[!t]
\centering
\caption{Comparison with state-of-the-art BNNs on ImageNet-1K. ``Search epochs" refers to the epochs required for both the training and searching steps to find the optimal architectures in automatically designed cases. ``Train/Finetune epochs" refers to the epochs required for either training or finetuning steps to train specialized models. ``$N$" indicates the number of deployment scenarios. ``\#100" indicates that the subnet is finetuned with 100 epochs after inheriting weights from the supernet.}
\label{tab:results}
\vspace{10pt}
\small
\resizebox{\linewidth}{!}{
\renewcommand\arraystretch{0.62}
\begin{tabular}{lrrrrrrr}
\toprule
\multirow{2}{*}{Model}   & \multirow{2}{*}{Design type}        & \multirow{2}{*}{OPs (M)} & \multirow{2}{*}{Top-1 (\%)} & Search       & \multicolumn{3}{c}{Train/Finetune epochs}    \\
                         &           &                          &                             & epochs       & $N$=1          & $N$=10          & $N$=40          \\
\midrule
\textbf{NAS-BNN-A}       & \textbf{auto} & \textbf{21}              & \textbf{60.58}              & \textbf{512} & \textbf{0}   & \textbf{0}    & \textbf{0}    \\
\textbf{NAS-BNN-A \#100} & \textbf{auto} & \textbf{21}              & \textbf{60.80}                   & \textbf{512} & \textbf{100} & \textbf{1000} & \textbf{4000} \\
Binary MobileNet-M3~\cite{phan2020binarizing}      & auto          & 33                       & 51.06                       & 64           & 256          & 2560          & 10240         \\
\textcolor{black}{MoBiNet-Mid (K=4)}~\cite{phan2020mobinet}       & \textcolor{black}{manual}        & \textcolor{black}{52}                       & \textcolor{black}{54.40}                       & \textcolor{black}{-}            & \textcolor{black}{50}           & \textcolor{black}{500}           & \textcolor{black}{2000}          \\
\midrule
\textbf{NAS-BNN-B}       & \textbf{auto} & \textbf{57}              & \textbf{66.46}              & \textbf{512} & \textbf{0}   & \textbf{0}    & \textbf{0}    \\
\textbf{NAS-BNN-B \#100} & \textbf{auto} & \textbf{57}              & \textbf{68.20}              & \textbf{512} & \textbf{100} & \textbf{1000} & \textbf{4000} \\
Binary MobileNet-M2~\cite{phan2020binarizing}      & auto          & 62                       & 59.30                       & 64           & 256          & 2560          & 10240         \\
\midrule
\textbf{NAS-BNN-C}       & \textbf{auto} & \textbf{87}              & \textbf{67.97}              & \textbf{512} & \textbf{0}   & \textbf{0}    & \textbf{0}    \\
\textbf{NAS-BNN-C \#100} & \textbf{auto} & \textbf{87}              & \textbf{69.49}                   & \textbf{512} & \textbf{100} & \textbf{1000} & \textbf{4000} \\
\textcolor{black}{ReActNet-A}~\cite{liu2020reactnet}               & \textcolor{black}{manual}        & \textcolor{black}{87}                       & \textcolor{black}{69.40}                       & \textcolor{black}{-}            & \textcolor{black}{512}          & \textcolor{black}{5120}          & \textcolor{black}{20480}         \\
BATS~\cite{bulat2020bats}                     & auto          & 99                       & 60.40                       & 25           & 75           & 750           & 3000          \\
\midrule
\textbf{NAS-BNN-D}       & \textbf{auto} & \textbf{124}              & \textbf{69.02}              & \textbf{512} & \textbf{0}   & \textbf{0}    & \textbf{0}    \\
\textbf{NAS-BNN-D \#100} & \textbf{auto} & \textbf{124}              & \textbf{70.26}                   & \textbf{512} & \textbf{100} & \textbf{1000} & \textbf{4000} \\
\textcolor{black}{BCNN P=1}~\cite{redfern2021bcnn}                 & \textcolor{black}{manual}        & \textcolor{black}{131}                      & \textcolor{black}{69.00}                       & \textcolor{black}{-}            & \textcolor{black}{130}          & \textcolor{black}{1300}          & \textcolor{black}{5200}          \\
\textcolor{black}{MeliusNet-C}~\cite{bethge2020meliusnet}              & \textcolor{black}{manual}        & \textcolor{black}{150}                      & \textcolor{black}{64.10}                       & \textcolor{black}{-}            & \textcolor{black}{120}          & \textcolor{black}{1200}          & \textcolor{black}{4800}          \\
Binary MobileNet-M1~\cite{phan2020binarizing}      & auto          & 154                      & 60.90                       & 64           & 256          & 2560          & 10240         \\
BATS (2x-wider)~\cite{bulat2020bats}          & auto          & 155                      & 66.10                       & 25           & 75           & 750           & 3000          \\
\midrule
\textbf{NAS-BNN-E}       & \textbf{auto} & \textbf{160}             & \textbf{69.48}              & \textbf{512} & \textbf{0}   & \textbf{0}    & \textbf{0}    \\
\textbf{NAS-BNN-E \#100} & \textbf{auto} & \textbf{160}             & \textbf{70.69}              & \textbf{512} & \textbf{100} & \textbf{1000} & \textbf{4000} \\
\textcolor{black}{MeliusNet-A}~\cite{bethge2020meliusnet}              & \textcolor{black}{manual}        & \textcolor{black}{162}                      & \textcolor{black}{63.40}                       & \textcolor{black}{-}            & \textcolor{black}{120}          & \textcolor{black}{1200}          & \textcolor{black}{4800}          \\
\textcolor{black}{Bi-RealNet-18}~\cite{liu2018bi}            & \textcolor{black}{manual}        & \textcolor{black}{163}                      & \textcolor{black}{56.40}                       & \textcolor{black}{-}           & \textcolor{black}{256}          & \textcolor{black}{2560}          & \textcolor{black}{10240}         \\
\textcolor{black}{DA-BNN}~\cite{zhao2022data}            & \textcolor{black}{manual}        & \textcolor{black}{163}                      & \textcolor{black}{66.30}                       & \textcolor{black}{-}           & \textcolor{black}{120}          & \textcolor{black}{1200}          & \textcolor{black}{4800}         \\
\textcolor{black}{ReActNet-B}~\cite{liu2020reactnet}               & \textcolor{black}{manual}        & \textcolor{black}{163}                      & \textcolor{black}{70.10}                       & \textcolor{black}{-}            & \textcolor{black}{512}          & \textcolor{black}{5120}          & \textcolor{black}{20480}         \\
BNAS-E~\cite{kim2020learning}                   & auto          & 163                      & 58.76                       & 50           & 250          & 2500          & 10000         \\
\textcolor{black}{XNOR-Net}~\cite{rastegari2016xnor}              & \textcolor{black}{manual}        & \textcolor{black}{167}                      & \textcolor{black}{51.20}                       & \textcolor{black}{-}           & \textcolor{black}{58}          & \textcolor{black}{580}          & \textcolor{black}{2320}          \\
\midrule
\textbf{NAS-BNN-F}       & \textbf{auto} & \textbf{180}             & \textbf{69.54}              & \textbf{512} & \textbf{0}   & \textbf{0}    & \textbf{0}    \\
\textbf{NAS-BNN-F \#100} & \textbf{auto} & \textbf{180}             & \textbf{70.80}              & \textbf{512} & \textbf{100} & \textbf{1000} & \textbf{4000} \\
\textcolor{black}{Real-to-Bin}~\cite{martinez2020training}              & \textcolor{black}{manual}        & \textcolor{black}{183}                      & \textcolor{black}{65.40}                       & \textcolor{black}{-}           & \textcolor{black}{150}          & \textcolor{black}{1500}          & \textcolor{black}{6000}          \\
\textcolor{black}{MeliusNet-B}~\cite{bethge2020meliusnet}              & \textcolor{black}{manual}        & \textcolor{black}{196}                      & \textcolor{black}{65.70}                       & \textcolor{black}{-}            & \textcolor{black}{120}          & \textcolor{black}{1200}          & \textcolor{black}{4800}         \\
\bottomrule
\end{tabular}
}
\end{table*}

%% file: sections/tables/results_det.tex
\begin{table}[!t]
    \centering
    \caption{Comparison with state-of-the-art binary detectors on MS COCO.}
    \vspace{10pt}
\renewcommand\arraystretch{1}
    \label{tab:results_det}
    \small
\begin{tabular}{llrr}
\toprule
Detector                                            & Method & OPs (M)      & mAP (\%)      \\ 
\midrule
\multicolumn{1}{c|}{\multirow{17}{*}{Faster R-CNN}} & \textbf{NAS-BNN-A}        & \textbf{140} & \textbf{25.0} \\
\multicolumn{1}{c|}{}                               & \textbf{NAS-BNN-B}        & \textbf{217} & \textbf{29.3} \\
\multicolumn{1}{c|}{}                               & \textbf{NAS-BNN-C}        & \textbf{279} & \textbf{30.6} \\
\cline{2-4}
\multicolumn{1}{c|}{}                               & \textbf{NAS-BNN-D}        & \textbf{370} & \textbf{31.6} \\
\multicolumn{1}{c|}{}                               & Bi-Real Net-18~\cite{liu2018bi}            & 376          & 17.4          \\
\multicolumn{1}{c|}{}                               & BiDet-18~\cite{wang2020bidet}                  & 376          & 19.4          \\
\multicolumn{1}{c|}{}                               & ReActNet-18~\cite{liu2020reactnet}               & 376          & 21.1          \\
\multicolumn{1}{c|}{}                               & LWS-Det-18~\cite{xu2021layer}                & 376          & 26.9          \\
\multicolumn{1}{c|}{}                               & Bi-Real Net-34~\cite{liu2018bi}            & 436          & 20.1          \\
\multicolumn{1}{c|}{}                               & BiDet-34~\cite{wang2020bidet}                  & 436          & 21.7          \\
\multicolumn{1}{c|}{}                               & ReActNet-34~\cite{liu2020reactnet}               & 436          & 23.4          \\
\multicolumn{1}{c|}{}                               & LWS-Det-34~\cite{xu2021layer}                & 436          & 29.9          \\ \cline{2-4}
\multicolumn{1}{c|}{}                               & \textbf{NAS-BNN-E}        & \textbf{443} & \textbf{32.1} \\
\multicolumn{1}{c|}{}                               & Bi-Real Net-50~\cite{liu2018bi}            & 459          & 22.9          \\
\multicolumn{1}{c|}{}                               & ReActNet-50~\cite{liu2020reactnet}               & 459          & 26.1          \\
\multicolumn{1}{c|}{}                               & LWS-Det-50~\cite{xu2021layer}                & 459          & 31.7          \\ \cline{2-4}
\multicolumn{1}{c|}{}                               & \textbf{NAS-BNN-F}        & \textbf{484} & \textbf{32.2} \\ 
\bottomrule
\end{tabular}
\end{table}

%% file: sections/tables/template.tex
\begin{table}[!t]
\centering
\caption{Comparison with different binary model templates in top-1 accuracy on ImageNet-100.}
\vspace{10pt}
\label{tab:template}
\small
\begin{tabular}{lrr}
\toprule
Model       & OPs~(M) & Top-1~(\%) \\
\midrule
Bi-Real-MobileNetV2~\cite{sandler2018mobilenetv2} & 97.48  & 60.84                  \\
Bi-Real-ResNet50~\cite{he2016deep}   & 174.62 & 67.46                  \\
\midrule
Bi-Real-MobileNetV1~\cite{howard2017mobilenets} & \textbf{86.20}   & \textbf{75.80}                  \\
\bottomrule
\end{tabular}
\end{table}

%% file: sections/tables/finetune.tex
\begin{table}[!t]
\centering
\caption{Effect of the finetuning step on pipeline 2. We evaluate the top-1 accuracy of NAS-BNN-B with different finetune epochs on ImageNet-1K.}
\vspace{10pt}
\label{tab:finetune}
\small
\begin{tabular}{lrrrrr}
\toprule
Finetune epochs & 0     & 25    & 50    & 75    & 100   \\
\midrule
Top-1~(\%)      & 66.46 & 67.67 & 67.92 & 68.04 & 68.20 \\
$\Delta~(\%)$   & -     & +1.21  & +1.46  & +1.58  & +1.74  \\
\bottomrule
\end{tabular}
\end{table}

%% file: sections/tables/arch.tex
\begin{table}[t]
    \centering
    \caption{\modified{Architecture summary of our specialized NAS-BNN models. S*-L*: ``S'' denotes the stage number and ``L'' denotes the layer number. c*-k*-g*: ``c'' denotes the channel width, ``k'' denotes the kernel size, and ``g'' denotes the number of groups.}}
    \label{tab:arch}
    \small
    \resizebox{\linewidth}{!}{
    \renewcommand\arraystretch{0.9}
\begin{tabular}{llrrrrrr}
\toprule
Input                      & Operator & A (21M OPs)             & B (57M OPs)              & C (87M OPs)             & D (124M OPs)            & E (160M OPs)            & F (180M OPs)             \\
\midrule
224 $\times$ 224                  & Conv2d   & c24\_k3\_g1   & c24\_k3\_g1    & c24\_k3\_g1    & c48\_k3\_g1   & c48\_k3\_g1   & c48\_k3\_g1    \\
\midrule
\multirow{3}{*}{112 $\times$ 112} & S1-L1    & c48\_k3\_g1   & c48\_k3\_g1    & c48\_k3\_g1    & c48\_k3\_g1   & c48\_k3\_g1   & c48\_k3\_g1    \\
                           & S1-L2    & c48\_k3\_g1   & c48\_k3\_g1    & c64\_k3\_g1    & c96\_k3\_g1   & c96\_k3\_g1   & c96\_k3\_g1    \\
                           & S1-L3    & -             & -              & -              & -             & -             & c96\_k3\_g1    \\
\midrule
\multirow{3}{*}{112 $\times$ 112} & S2-L1    & c96\_k3\_g2   & c96\_k3\_g2    & c96\_k5\_g2    & c192\_k5\_g2  & c192\_k5\_g2  & c192\_k3\_g1   \\
                           & S2-L2    & c96\_k3\_g2   & c192\_k3\_g2   & c192\_k3\_g1   & c192\_k3\_g2  & c192\_k3\_g2  & c192\_k3\_g1   \\
                           & S2-L3    & -             & -              & -              & -             & -             & -              \\
\midrule
\multirow{3}{*}{56 $\times$ 56}   & S3-L1    & c192\_k3\_g4  & c192\_k3\_g2   & c384\_k3\_g2   & c384\_k3\_g2  & c384\_k3\_g2  & c384\_k5\_g2   \\
                           & S3-L2    & c192\_k3\_g4  & c384\_k3\_g2   & c384\_k3\_g4   & c384\_k3\_g4  & c384\_k3\_g4  & c384\_k3\_g2   \\
                           & S3-L3    & -             & -              & c384\_k3\_g4   & c384\_k3\_g4  & c384\_k3\_g4  & c384\_k3\_g2   \\
\midrule
\multirow{9}{*}{28 $\times$ 28}   & S4-L1    & c384\_k3\_g8  & c384\_k5\_g8   & c768\_k5\_g4   & c768\_k3\_g4  & c768\_k3\_g4  & c768\_k5\_g4   \\
                           & S4-L2    & c384\_k3\_g8  & c768\_k3\_g8   & c768\_k3\_g4   & c768\_k3\_g4  & c768\_k3\_g4  & c768\_k5\_g4   \\
                           & S4-L3    & c384\_k3\_g8  & c768\_k3\_g8   & c768\_k3\_g8   & c768\_k3\_g4  & c768\_k3\_g4  & c768\_k5\_g4   \\
                           & S4-L4    & c384\_k3\_g8  & c768\_k3\_g8   & c768\_k3\_g4   & c768\_k5\_g8  & c768\_k5\_g8  & c768\_k3\_g4   \\
                           & S4-L5    & c384\_k3\_g8  & c768\_k3\_g8   & c768\_k3\_g4   & c768\_k5\_g4  & c768\_k5\_g4  & c768\_k3\_g4   \\
                           & S4-L6    & c384\_k3\_g8  & c768\_k3\_g4   & c768\_k3\_g8   & c768\_k3\_g4  & c768\_k3\_g4  & c768\_k3\_g4   \\
                           & S4-L7    & c384\_k3\_g8  & c768\_k3\_g4   & c768\_k5\_g8   & c768\_k5\_g4  & c768\_k5\_g4  & c768\_k5\_g4   \\
                           & S4-L8    & c384\_k3\_g8  & c768\_k3\_g8   & c768\_k3\_g8   & c768\_k5\_g8  & c768\_k5\_g8  & c768\_k5\_g4   \\
                           & S4-L9    & -             & -              & c768\_k3\_g4   & c768\_k3\_g4  & c768\_k3\_g4  & c768\_k5\_g4   \\
\midrule
\multirow{3}{*}{14 $\times$ 14}   & S5-L1    & c768\_k3\_g16 & c1536\_k5\_g16 & c1536\_k3\_g8  & c1536\_k5\_g8 & c1536\_k5\_g8 & c1536\_k3\_g16 \\
                           & S5-L2    & c768\_k3\_g16 & c1536\_k5\_g16 & c1536\_k5\_g16 & c1536\_k5\_g8 & c1536\_k5\_g8 & c1536\_k5\_g8  \\
                           & S5-L3    & -             & c1536\_k3\_g16 & c1536\_k3\_g8  & c1536\_k3\_g8 & c1536\_k3\_g8 & c1536\_k3\_g8 \\
\bottomrule
\end{tabular}}
    \end{table}

%% file: sections/5_conclusion.tex
\section{Conclusion}
In this paper, we present a novel neural architecture search scheme specifically designed for binary neural networks, named NAS-BNN.
Specifically, we first carefully design a search space that is binary-friendly, including the traditional configurations and the specific configurations for BNNs. In addition, we propose three simple yet efficient training techniques to enhance the training process of binary supernet, including Bi-Teacher to supervise the training of subnets, Bi-Transformation, and Weight Normalization to enhance the representation ability of the shared weights.
In the end, we investigate the balance of supernet training and subnet finetuning and propose two deployment pipelines.
Our experimental results demonstrate that the searched binary models outperform previous BNNs in terms of both operations~(OPs) and accuracy \modified{on both image classification and object detection tasks}.

%% file: sections/6_reference.tex
{
\bibliographystyle{elsarticle-num} 
\bibliography{refs}
}

%% file: main.bbl
\begin{thebibliography}{10}
\expandafter\ifx\csname url\endcsname\relax
  \def\url#1{\texttt{#1}}\fi
\expandafter\ifx\csname urlprefix\endcsname\relax\def\urlprefix{URL }\fi
\expandafter\ifx\csname href\endcsname\relax
  \def\href#1#2{#2} \def\path#1{#1}\fi

\bibitem{qin2020binary}
H.~Qin, R.~Gong, X.~Liu, X.~Bai, J.~Song, N.~Sebe, Binary neural networks: A survey, Pattern Recognition 105 (2020) 107281.

\bibitem{hubara2016binarized}
I.~Hubara, M.~Courbariaux, D.~Soudry, R.~El-Yaniv, Y.~Bengio, Binarized neural networks, {Advances in Neural Information Processing Systems} 29 (2016).

\bibitem{liu2018bi}
Z.~Liu, B.~Wu, W.~Luo, X.~Yang, W.~Liu, K.-T. Cheng, Bi-real net: Enhancing the performance of 1-bit cnns with improved representational capability and advanced training algorithm, in: {European Conference on Computer Vision}, 2018, pp. 722--737.

\bibitem{ignatov2020controlling}
D.~Ignatov, A.~Ignatov, Controlling information capacity of binary neural network, Pattern Recognition Letters 138 (2020) 276--281.

\bibitem{phan2020mobinet}
H.~Phan, Y.~He, M.~Savvides, Z.~Shen, et~al., Mobinet: A mobile binary network for image classification, in: {IEEE/CVF Winter Conference on Applications of Computer Vision}, 2020, pp. 3453--3462.

\bibitem{liu2020reactnet}
Z.~Liu, Z.~Shen, M.~Savvides, K.-T. Cheng, Reactnet: Towards precise binary neural network with generalized activation functions, in: {European Conference on Computer Vision}, Springer, 2020, pp. 143--159.

\bibitem{peng2019bdnn}
H.~Peng, S.~Chen, Bdnn: Binary convolution neural networks for fast object detection, Pattern Recognition Letters 125 (2019) 91--97.

\bibitem{wang2020bidet}
Z.~Wang, Z.~Wu, J.~Lu, J.~Zhou, Bidet: An efficient binarized object detector, in: {IEEE/CVF Conference on Computer Vision and Pattern Recognition}, 2020, pp. 2049--2058.

\bibitem{zhao2022data}
J.~Zhao, S.~Xu, R.~Wang, B.~Zhang, G.~Guo, D.~Doermann, D.~Sun, Data-adaptive binary neural networks for efficient object detection and recognition, Pattern Recognition Letters 153 (2022) 239--245.

\bibitem{xu2021layer}
S.~Xu, J.~Zhao, J.~Lu, B.~Zhang, S.~Han, D.~Doermann, Layer-wise searching for 1-bit detectors, in: {IEEE/CVF Conference on Computer Vision and Pattern Recognition}, 2021, pp. 5682--5691.

\bibitem{qin2020bipointnet}
H.~Qin, Z.~Cai, M.~Zhang, Y.~Ding, H.~Zhao, S.~Yi, X.~Liu, H.~Su, Bipointnet: Binary neural network for point clouds, in: {International Conference on Learning Representations}, 2020, pp. 1--24.

\bibitem{rastegari2016xnor}
M.~Rastegari, V.~Ordonez, J.~Redmon, A.~Farhadi, Xnor-net: Imagenet classification using binary convolutional neural networks, in: {European Conference on Computer Vision}, Springer, 2016, pp. 525--542.

\bibitem{jiang2019bitstream}
Y.~Jiang, T.~Zhao, X.~He, C.~Leng, J.~Cheng, Bitstream: An efficient framework for inference of binary neural networks on cpus, Pattern Recognition Letters 125 (2019) 303--309.

\bibitem{kim2020learning}
D.~Kim, K.~P. Singh, J.~Choi, Learning architectures for binary networks, in: {European Conference on Computer Vision}, Springer, 2020, pp. 575--591.

\bibitem{bulat2020bats}
A.~Bulat, B.~Martinez, G.~Tzimiropoulos, Bats: Binary architecture search, in: {European Conference on Computer Vision}, Springer, 2020, pp. 309--325.

\bibitem{zhu2020nasb}
B.~Zhu, Z.~Al-Ars, H.~P. Hofstee, Nasb: Neural architecture search for binary convolutional neural networks, in: {IEEE International Joint Conference on Neural Networks}, IEEE, 2020, pp. 1--8.

\bibitem{zhao2020bars}
T.~Zhao, X.~Ning, X.~Shi, S.~Yang, S.~Liang, P.~Lei, J.~Chen, H.~Yang, Y.~Wang, Bars: Joint search of cell topology and layout for accurate and efficient binary architectures, arXiv preprint arXiv:2011.10804 (2020).

\bibitem{phan2020binarizing}
H.~Phan, Z.~Liu, D.~Huynh, M.~Savvides, K.-T. Cheng, Z.~Shen, Binarizing mobilenet via evolution-based searching, in: {IEEE/CVF Conference on Computer Vision and Pattern Recognition}, 2020, pp. 13420--13429.

\bibitem{liu2018darts}
H.~Liu, K.~Simonyan, Y.~Yang, Darts: Differentiable architecture search, in: {International Conference on Learning Representations}, 2019, pp. 1--13.

\bibitem{cai2019once}
H.~Cai, C.~Gan, T.~Wang, Z.~Zhang, S.~Han, Once-for-all: Train one network and specialize it for efficient deployment, in: {International Conference on Learning Representations}, 2020, pp. 1--15.

\bibitem{yu2019universally}
J.~Yu, T.~S. Huang, Universally slimmable networks and improved training techniques, in: {IEEE/CVF International Conference on Computer Vision}, 2019, pp. 1803--1811.

\bibitem{yu2020bignas}
J.~Yu, P.~Jin, H.~Liu, G.~Bender, P.-J. Kindermans, M.~Tan, T.~Huang, X.~Song, R.~Pang, Q.~Le, Bignas: Scaling up neural architecture search with big single-stage models, in: {European Conference on Computer Vision}, Springer, 2020, pp. 702--717.

\bibitem{wang2021attentivenas}
D.~Wang, M.~Li, C.~Gong, V.~Chandra, Attentivenas: Improving neural architecture search via attentive sampling, in: {IEEE/CVF Conference on Computer Vision and Pattern Recognition}, 2021, pp. 6418--6427.

\bibitem{wang2021alphanet}
D.~Wang, C.~Gong, M.~Li, Q.~Liu, V.~Chandra, Alphanet: Improved training of supernets with alpha-divergence, in: {International Conference on Machine Learning}, PMLR, 2021, pp. 10760--10771.

\bibitem{hu2022learning}
Y.~Hu, N.~Belkhir, J.~Angulo, A.~Yao, G.~Franchi, Learning deep morphological networks with neural architecture search, Pattern Recognition 131 (2022) 108893.

\bibitem{tong2022neural}
L.~Tong, B.~Du, Neural architecture search via reference point based multi-objective evolutionary algorithm, Pattern Recognition 132 (2022) 108962.

\bibitem{he2016deep}
K.~He, X.~Zhang, S.~Ren, J.~Sun, Deep residual learning for image recognition, in: {IEEE/CVF Conference on Computer Vision and Pattern Recognition}, 2016, pp. 770--778.

\bibitem{howard2017mobilenets}
A.~G. Howard, M.~Zhu, B.~Chen, D.~Kalenichenko, W.~Wang, T.~Weyand, M.~Andreetto, H.~Adam, Mobilenets: Efficient convolutional neural networks for mobile vision applications, arXiv preprint arXiv:1704.04861 (2017).

\bibitem{bulat2019improved}
A.~Bulat, G.~Tzimiropoulos, J.~Kossaifi, M.~Pantic, Improved training of binary networks for human pose estimation and image recognition, arXiv preprint arXiv:1904.05868 (2019).

\bibitem{martinez2020training}
B.~Martinez, J.~Yang, A.~Bulat, G.~Tzimiropoulos, Training binary neural networks with real-to-binary convolutions, in: {International Conference on Learning Representations}, 2020, pp. 1--11.

\bibitem{bethge2020meliusnet}
J.~Bethge, C.~Bartz, H.~Yang, Y.~Chen, C.~Meinel, Meliusnet: Can binary neural networks achieve mobilenet-level accuracy?, arXiv preprint arXiv:2001.05936 (2020).

\bibitem{esser2019learned}
S.~K. Esser, J.~L. McKinstry, D.~Bablani, R.~Appuswamy, D.~S. Modha, Learned step size quantization, in: {International Conference on Learning Representations}, 2020, pp. 1--12.

\bibitem{lin2022fq}
Y.~Lin, T.~Zhang, P.~Sun, Z.~Li, S.~Zhou, Fq-vit: Post-training quantization for fully quantized vision transformer, in: Proceedings of the Thirty-First International Joint Conference on Artificial Intelligence, IJCAI-22, 2022, pp. 1173--1179.

\bibitem{zoph2016neural}
B.~Zoph, Q.~V. Le, Neural architecture search with reinforcement learning, in: {International Conference on Learning Representations}, 2017, pp. 1--16.

\bibitem{liu2017hierarchical}
H.~Liu, K.~Simonyan, O.~Vinyals, C.~Fernando, K.~Kavukcuoglu, Hierarchical representations for efficient architecture search, in: {International Conference on Learning Representations}, 2018, pp. 1--13.

\bibitem{zoph2018learning}
B.~Zoph, V.~Vasudevan, J.~Shlens, Q.~V. Le, Learning transferable architectures for scalable image recognition, in: {IEEE/CVF Conference on Computer Vision and Pattern Recognition}, 2018, pp. 8697--8710.

\bibitem{real2019regularized}
E.~Real, A.~Aggarwal, Y.~Huang, Q.~V. Le, Regularized evolution for image classifier architecture search, in: {Association for the Advancement of Artificial Intelligence}, 2019, pp. 4780--4789.

\bibitem{brock2017smash}
A.~Brock, T.~Lim, J.~M. Ritchie, N.~Weston, Smash: one-shot model architecture search through hypernetworks, in: {International Conference on Learning Representations}, 2018, pp. 1--22.

\bibitem{pham2018efficient}
H.~Pham, M.~Guan, B.~Zoph, Q.~Le, J.~Dean, Efficient neural architecture search via parameters sharing, in: {International Conference on Machine Learning}, PMLR, 2018, pp. 4095--4104.

\bibitem{cai2018proxylessnas}
H.~Cai, L.~Zhu, S.~Han, Proxylessnas: Direct neural architecture search on target task and hardware, in: {International Conference on Learning Representations}, 2019, pp. 1--13.

\bibitem{wu2019fbnet}
B.~Wu, X.~Dai, P.~Zhang, Y.~Wang, F.~Sun, Y.~Wu, Y.~Tian, P.~Vajda, Y.~Jia, K.~Keutzer, Fbnet: Hardware-aware efficient convnet design via differentiable neural architecture search, in: {IEEE/CVF Conference on Computer Vision and Pattern Recognition}, 2019, pp. 10734--10742.

\bibitem{guo2020single}
Z.~Guo, X.~Zhang, H.~Mu, W.~Heng, Z.~Liu, Y.~Wei, J.~Sun, Single path one-shot neural architecture search with uniform sampling, in: {European Conference on Computer Vision}, Springer, 2020, pp. 544--560.

\bibitem{wang2020apq}
T.~Wang, K.~Wang, H.~Cai, J.~Lin, Z.~Liu, H.~Wang, Y.~Lin, S.~Han, Apq: Joint search for network architecture, pruning and quantization policy, in: {IEEE/CVF Conference on Computer Vision and Pattern Recognition}, 2020, pp. 2078--2087.

\bibitem{bai2021batchquant}
H.~Bai, M.~Cao, P.~Huang, J.~Shan, Batchquant: Quantized-for-all architecture search with robust quantizer, {Advances in Neural Information Processing Systems} 34 (2021) 1074--1085.

\bibitem{sandler2018mobilenetv2}
M.~Sandler, A.~Howard, M.~Zhu, A.~Zhmoginov, L.-C. Chen, Mobilenetv2: Inverted residuals and linear bottlenecks, in: {IEEE/CVF Conference on Computer Vision and Pattern Recognition}, 2018, pp. 4510--4520.

\bibitem{Li2020Additive}
Y.~Li, X.~Dong, W.~Wang, Additive powers-of-two quantization: An efficient non-uniform discretization for neural networks, in: {International Conference on Learning Representations}, 2020, pp. 1--15.

\bibitem{qin2020forward}
H.~Qin, R.~Gong, X.~Liu, M.~Shen, Z.~Wei, F.~Yu, J.~Song, Forward and backward information retention for accurate binary neural networks, in: {IEEE/CVF Conference on Computer Vision and Pattern Recognition}, 2020, pp. 2250--2259.

\bibitem{redfern2021bcnn}
A.~J. Redfern, L.~Zhu, M.~K. Newquist, Bcnn: A binary cnn with all matrix ops quantized to 1 bit precision, in: {IEEE/CVF Conference on Computer Vision and Pattern Recognition}, 2021, pp. 4604--4612.

\bibitem{wu2018mixed}
B.~Wu, Y.~Wang, P.~Zhang, Y.~Tian, P.~Vajda, K.~Keutzer, Mixed precision quantization of convnets via differentiable neural architecture search, arXiv preprint arXiv:1812.00090 (2018).

\bibitem{koryakovskiy2023one}
I.~Koryakovskiy, A.~Yakovleva, V.~Buchnev, T.~Isaev, G.~Odinokikh, One-shot model for mixed-precision quantization, in: {IEEE/CVF Conference on Computer Vision and Pattern Recognition}, 2023, pp. 7939--7949.

\end{thebibliography}
